\documentclass[final]{cvpr}

\usepackage{times}
\usepackage{epsfig}
\usepackage{graphicx}
\usepackage{amsmath}
\usepackage{amssymb}
\usepackage{amsmath}
\usepackage{multirow}
\newcommand{\minitab}[2][c]{\begin{tabular}{#1}#2\end{tabular}}

\usepackage[pagebackref=true,breaklinks=true,colorlinks,bookmarks=false]{hyperref}

\begin{document}

\title{FBC-GAN: Diverse and Flexible Image Synthesis via\\ Foreground-Background Composition}

\author{
Kaiwen Cui  \qquad Gongjie Zhang \qquad Fangneng Zhan\qquad Jiaxing Huang\qquad Shijian Lu \thanks{corresponding author} \smallskip\\
{Nanyang Technological University, Singapore}\\
{\tt\footnotesize kaiwen001@e.ntu.edu.sg  \quad  shijian.lu@ntu.edu.sg}
}

\maketitle

\begin{abstract}

Generative Adversarial Networks (GANs) have become the de-facto standard in image synthesis. However, without considering the foreground-background decomposition, existing GANs tend to capture excessive content correlation between foreground and background, thus constraining the diversity in image generation. This paper presents a novel Foreground-Background Composition GAN (FBC-GAN) that performs image generation by generating foreground objects and background scenes concurrently and independently, followed by composing them with style and geometrical consistency. With this explicit design, FBC-GAN can generate images with foregrounds and backgrounds that are mutually independent in contents, thus lifting the undesirably learnt content correlation constraint and achieving superior diversity. It also provides excellent flexibility by allowing the same foreground object with different background scenes, the same background scene with varying foreground objects, or the same foreground object and background scene with different object positions, sizes and poses. It can compose foreground objects and background scenes sampled from different datasets as well.
Extensive experiments over multiple datasets show that FBC-GAN achieves competitive visual realism and superior diversity as compared with state-of-the-art methods.

\end{abstract}

\section{Introduction}  \label{sec:1}

\begin{figure}[t!] 
\begin{center}
   \includegraphics[width=1\linewidth, height=90mm]{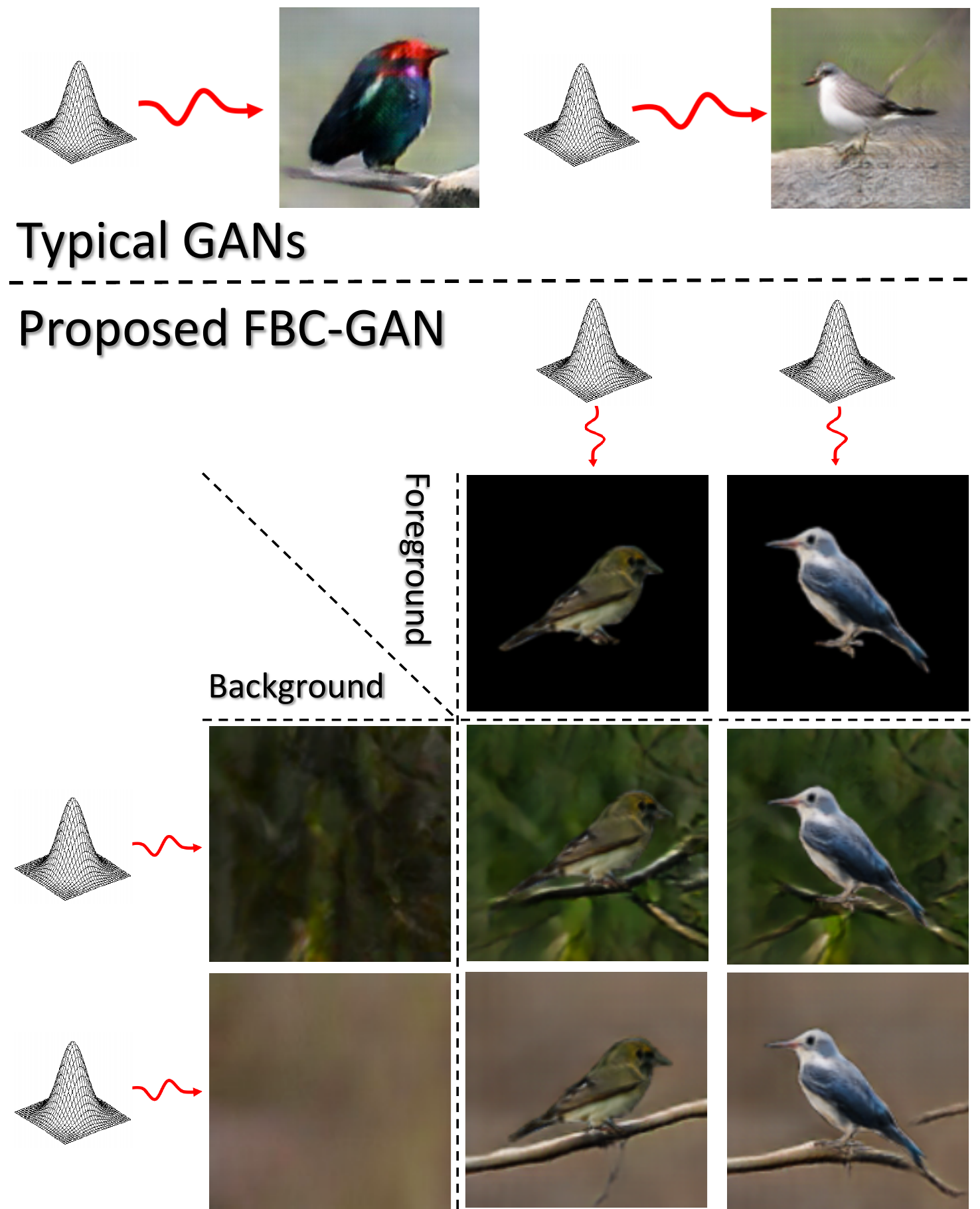}
\end{center}
   \caption{
   Illustration of the proposed FBC-GAN: \textit{Typical GANs} synthesize images in a direct and single-step manner by using one randomly sampled latent code, which tends to suffer from constrained diversity and flexibility in generated contents. FBC-GAN offers much richer diversity and flexibility by first synthesizing style-compatible background scenes and foreground objects independently and then composing them with geometrical plausibility.
   }
\label{fig:1}
\end{figure}

Imagine this scene: a sparrow is standing on a branch, and it flies away and lands on another branch. A while later, the sparrow flies out of sight, and another cuckoo lands on this very branch to take a rest from a long flight. It is natural that one object can appear in different scenes, and so can various distinct objects appear in the same scene. Obviously, most images can be typically decomposed into one or multiple foreground objects and a background scene.

Ideally, image synthesis should be equipped with such capability for capturing the diversity of foreground-background composition. On the other hand, most existing GANs are `Typical GANs’ as illustrated in Fig.~\ref{fig:1}, which generate new images at one go without separate handling of foreground objects and background scenes. Consequently, the generated images usually have highly correlated foregrounds and backgrounds depending on the training images. We argue that the foreground-background correlation usually consists of three components including 1) \textit{style correlation} that captures consistency in color, saturation, etc., 2) \textit{geometry correlation} that captures geometrical plausibility,
and 3) \textit{content correlation} that captures the co-occurrence of foreground objects and background scenes\,\cite{zhang2019cad}. `Typical GANs’ tend to capture excessive \textit{content correlation}, especially when the training data do not have sufficient size and variations, leading to limited diversity in the generated images.

Some attempts have been reported to address the excessive content correlation by generating foregrounds and backgrounds separately. One work is LR-GAN~\cite{yang2017lr} which first generates backgrounds and foregrounds recursively and then stitches them to produce the final synthesis. However, LR-GAN essentially belongs to `Typical GANs’ as its generated foreground is entirely conditioned on the generated background and the whole generation still conditions on a single latent code. Another work is FineGAN~\cite{zhao2019image} which disentangles backgrounds and foregrounds and generates them hierarchically. However, FineGAN cannot handle foreground-background mismatch well, and its generated foreground tends to occlude background completely when mismatches happen, making FineGAN degrade to the `Typical GANs'. Additionally, the stitching process in FineGAN does not guarantee style alignment, i.e. the generated foreground does not adjust its style for matching with the generated background. To the best of our knowledge, these are only two works that attempt to generate image foreground and background separately. 

This paper presents Foreground-Background Composition GAN (FBC-GAN) - an innovative image synthesis network that achieves superior synthesis flexibility and diversity by generating image foreground and background independently as illustrated in Fig.~\ref{fig:1}. This independent generation approach relaxes not only content correlation (undesired) but also style and geometry correlations (desired) between the generated foregrounds and backgrounds. To reinstate the style and geometry correlations, we adapt the idea of AdaIN~\cite{huang2017arbitrary} for style alignment and modify the generated background for geometric plausibility between the generated foregrounds and backgrounds. Leveraging these designs, FBC-GAN can offer superior diversity and flexibility in image synthesis, e.g., it can generate images with the same foreground object but different background scenes, the same background scene but different foreground objects, or the same foreground object and background scene but different object positions, sizes, and poses without additional conditions. Additionally, it allows generating `new information’ by enabling image synthesis with foreground and background sampled from different datasets.

The contributions of this work are threefold. \textit{First}, we propose FBC-GAN -- a novel image synthesis network that relaxes the excessive content correlation by generating image foreground and background independently. \textit{Second}, we design novel consistency mechanisms that achieve style alignment and geometry plausibility (between independently generated foreground and background) by exploiting feature statistics and adaptive translation of the generated image background. \textit{Third}, FBC-GAN enables image synthesis with foreground and background sampled from different datasets, which allows to generate rich and new information beyond the distribution of a single reference dataset as in most existing image synthesis GANs.

\section{Related Works}    \label{sec:2}

\textbf{Image Synthesis.} 
In addition to discriminative models\;\cite{zhang2019cad,zhang2021meta,zhang2021pnpdet,zhang2021detr,huang2021fsdr, huang2021rda}, generative models\,\cite{goodfellow2014generative} have achieved remarkable progress in recent years, especially in image synthesis. 
It finds applications in different tasks in image translation \cite{park2019semantic,Isola_2017_CVPR,zhan2021unite,zhan2020sagan,zhang2021defect,zhang2021crlsr,zhan2021rabit},
image inpainting~\cite{iizuka2017globally,yu2019free,yu2021diverse}, 
and image editing \cite{yu2018inpainting,wu2020cascade,wu2020leed,zhan2021emlight,zhan2021gmlight,zhan2020towards,zhan2021needlelight}.

The early researches focus on synthesizing images unconditionally \cite{dcgan,mao2017least,karras2017progressive}. To achieve better controllability over certain attributes of the generated images, more and more works emerge to perform conditional image synthesis. Various conditions have been exploited including image labels~\cite{CGAN}, input images~\cite{Pathak_2016_CVPR,Isola_2017_CVPR,liu2017unsupervised,cyclegan,zhao2018modular,wu2019relgan}, sketches~\cite{sangkloy2017scribbler,zhu2017toward,xian2018texturegan,zhao2019image}, text descriptions~\cite{zhang2017stackgan,zhang2018stackgan++,cha2019adversarial}, etc.

Although conditional GANs impose certain controls over the synthesized images, most of them suffer from constrained diversity and flexibility in the generated contents as they synthesize images at one go from a single latent code. LR-GAN~\cite{yang2017lr} and FineGAN~\cite{zhao2019image} attempt to mitigate this problem by decomposing foreground and background and generating them separately. They propose a stitching based approach that first generates background and then stitches a correlated foreground into the background. However, the stitching tends to occlude background if the generated foreground does not match the background.  Our FBC-GAN generates foreground and background independently and composes them into a style-consistent and geometry-consistent image. It offers great flexibility and diversity in image generation, more details to be described in the ensuing subsections.

\begin{figure*}[t!] 
\begin{center}
   \includegraphics[width=0.975\linewidth, height=80mm]{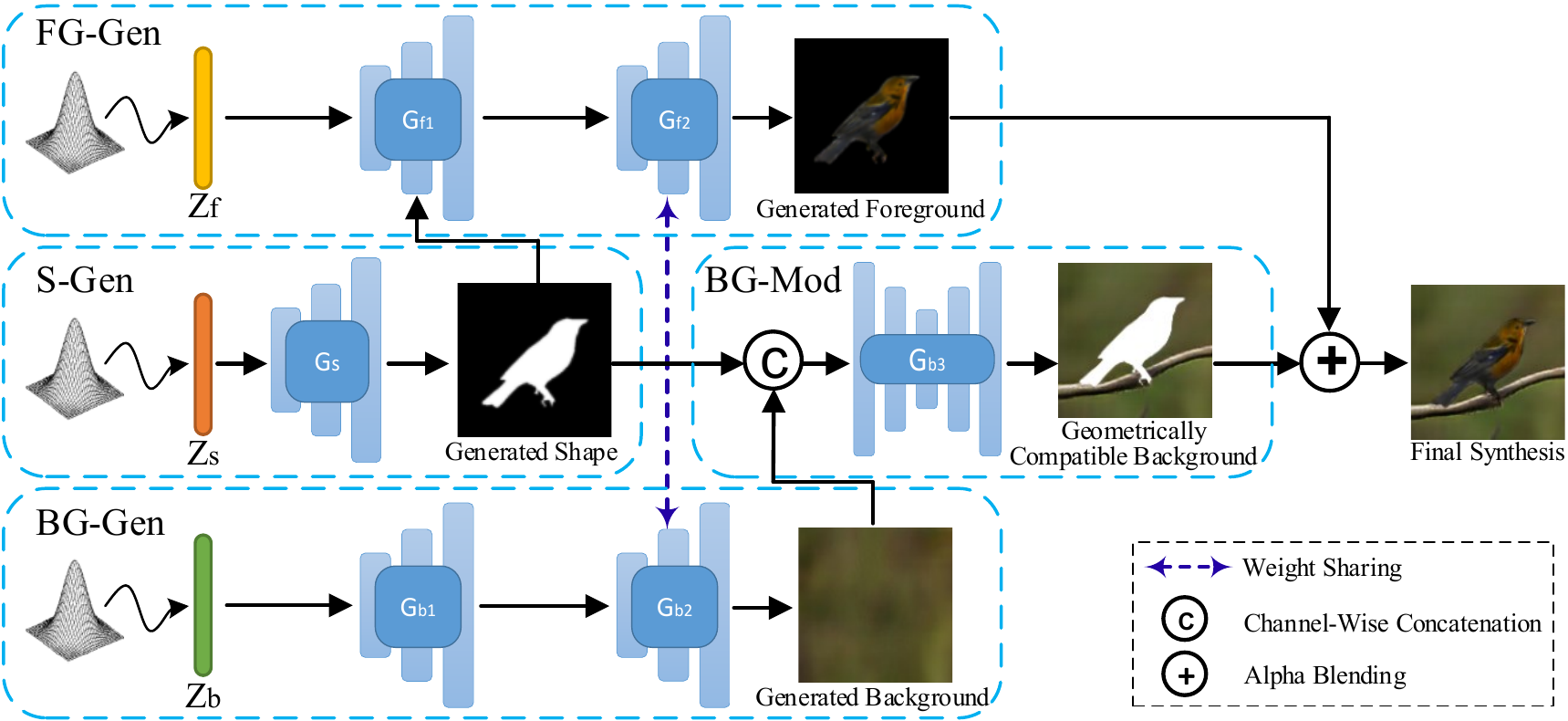}
\end{center}
   \caption{
   Architecture of the proposed FBC-GAN: FBC-GAN consists of a Shape Generator (S-Gen), a Foreground Generator (FG-Gen),  a Background Generator (BG-Gen) and a Background Modifier (BG-Mod). S-Gen first generates foreground shapes that are aims for bridging the geometry between the generated foreground and background. FG-Gen generates foreground objects that have the same shapes as the generated shapes. BG-Gen generates background scenes and BG-Mod modifies the generated backgrounds to be geometrically compatible for the generated shapes. Style alignment is achieved by sharing feature statistics of the generated backgrounds and foregrounds.
   }
\label{fig:model}
\end{figure*}

\smallskip

\noindent\textbf{Image Composition.} Image composition aims to overlay a masked foreground object over a background image with good consistency in both styles and geometry. To achieve style consistency, \cite{lalonde2007photo} selects foreground objects from a database with coherent appearance for composition. \cite{chen2019toward} uses a brightness and contrast model to adjust the foreground appearance. ~\cite{zhan2019spatial} incorporates CycleGAN~\cite{cyclegan} to align the foreground style with the background. For geometry consistency, most of the existing works~\cite{lin2018st,chen2019toward,zhan2019spatial,zhan2020aicnet,zhan2019esir,zhan2019gadan,zhan2018verisimilar,zhan2019scene} leverage a Spatial Transformation Network~\cite{jaderberg2015spatial} to adjust the position, size and pose of foreground objects.

Our FBC-GAN is composition based model which generates foreground and background separately. It adopts Adaptive Instance Normalization (AdaIN)~\cite{huang2017arbitrary} to achieve style consistency in the its generated foreground and background. For geometry consistency, it designs an innovative composition module that employs foreground shape as guidance to guide the generation of foreground and modify the background with minor alternations for geometrical compatibility.

\section{Method}

\subsection{Overview}  

The proposed FBC-GAN is an end-to-end image synthesizer. It consists of four modules: Shape Generator (S-Gen), Foreground Generator (FG-Gen), Background Generator (BG-Gen) and Background Modifier (BG-Mod) as illustrated in Fig.~\ref{fig:model}. FG-Gen and BG-Gen independently and concurrently generate image foregrounds and backgrounds from latent codes sampled from Gaussian distributions. By generating foreground and background separately, it lifts the content, style and geometry correlations between them. To reinstate the style correlation, it aligns the statistics of their feature representations as inspired by AdaIN~\cite{huang2017arbitrary}. To reinstate the geometry correlation, it employs a S-Gen and BG-Mod to bridge the geometry (e.g., adding a branch below a bird standing in the sky) between the independently generated foregrounds and backgrounds.

As a result, the proposed FBC-GAN retains the style and geometry consistency for generated images for visual realism, while lifts the content correlations between generated foregrounds and backgrounds for superior synthesis diversity and flexibility.

\subsection{Foreground Generation} \label{Foreground Generation}

\textit{Foreground Generator (FG-Gen)} is illustrated in the upper part of Fig.~\ref{fig:model}. 
The network architecture of FG-Gen follows the generator in SPADE~\cite{park2019semantic} except it takes random Gaussian latent codes as input and the shape generated by S-Gen serves as input to SPADE module to guide the foreground generation. We divide the generator into two parts, $G_{f1}$ and $G_{f2}$, and $G_{f2}$ is only a single convolution layer that shares weight with $G_{b2}$ in Background Generator.

Since FG-Gen and BG-Gen generate foreground and background separately and independently, the style of the generated foreground and background are usually incompatible with each other. Style alignment of foreground and background is thus necessary for synthesis realism. We adopt AdaIN~\cite{huang2017arbitrary} to feature map inputs of $G_{f2}$ and $G_{b2}$ to achieve style alignment of foreground and background. To align styles without changing the distinctive appearance of foreground, we modify AdaIN to soft AdaIN where style-aligned foreground features are weighted combination of features obtained from AdaIN and the original features: 
\begin{align}\begin{aligned}
F_{f\_sc}= \alpha  AdaIN(F_{f}, F_{b}) + (1 - \alpha)  F_{f}
\end{aligned}\end{align}
where $F_{f\_sc}$ is the style-aligned feature maps of foreground. $F_{f}$ and $F_{b}$ are feature map inputs of $G_{f2}$ and $G_{b2}$ respectively. $\alpha$ is a hyper-parameter controlling the intensity of the style alignment operation, which is empirically set to $0.2$ if foregrounds and backgrounds are all nature style images (Dataset1, Dataset2 and Dataset4 introduced in section \ref{dataset}) and $0.4$ if foregrounds are nature style and backgrounds are Monet style images (Dataset3 and Dataset5 introduced in section \ref{dataset}).

Finally, $F_{f\_sc}$ is fed into $G_{f2}$ to generate style-aligned foreground object. 
\begin{figure}[t!] 
\begin{center}
   \includegraphics[width=1.0\linewidth, height=64mm]{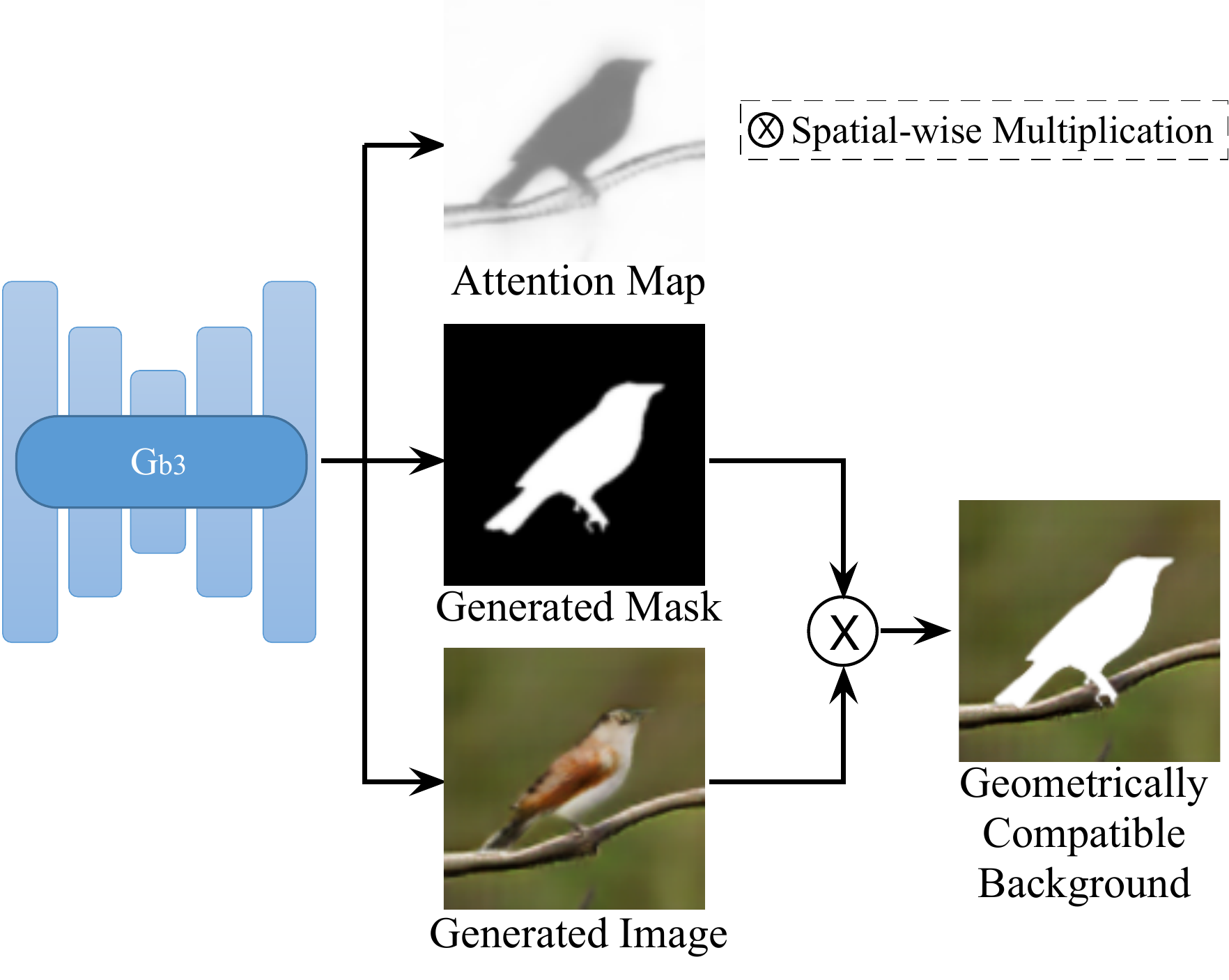}
\end{center}
\caption{Details of $G_{b3}$ in Fig.~\ref{fig:model}: $G_{b3}$ outputs an \textit{Attention Map}, a binary \textit{Generated Mask} and a \textit{Generated Image}, where pixel-wise multiplication of the later two produces a \textit{Geometrically Compatible Background}. The attention map is used during training to guide $G_{b3}$ to achieve geometrical consistency with least modification.}
\label{fig:model2}
\end{figure}

\subsection{Background Generation} 
Background Generation consists of two modules: \textit{Background Generator (BG-Gen}) and \textit{Background Modifier (BG-Mod}). 

As shown in Fig.~\ref{fig:model}, \textit{BG-Gen} mainly consists of two parts, $G_{b1}$ and $G_{b2}$.  $G_{b1}$ and $G_{b2}$ together generate background scenes with random Gaussian latent codes as input. $G_{b2}$ shares weight with $G_{f2}$ for style alignment mentioned in section~\ref{Foreground Generation}.

\textit{BG-Mod} modifies the generated background to fit the generated shape in a geometrically consistent manner with the least alternation, ensuring geometrical consistency while keeping the background unchanged as much as possible. It concatenates the shape generated by S-Gen and the background generated by BG-Gen and feeds the concatenated results into $G_{b3}$.  Without considering foreground appearance, $G_{b3}$ focuses on aligning foreground and background geometrically.
Details of BG-Mod is illustrated in Fig.~\ref{fig:model2}. The preliminary output of $G_{b3}$ is a geometrically aligned image, a foreground binary mask and an attention map. 
The geometrically aligned image is generated through adversarial learning. We use the idea of GAN-CLS~\cite{reed2016generative} to match the generated images and generated foreground binary masks so that the foreground objects shapes of generated images can be the same as the input shape. Spatial-wisely multiplying the generated mask and image produces geometrically compatible backgrounds as shown in Fig.~\ref{fig:model}. Attention map is used only during training stage to guide the generation of background contents. With generated shape as reference, it pays as much attention as possible to areas outside the foreground object (white pixels) and allows minor alternation  of the background (black pixels) to achieve geometry consistency.

\subsection{Final Image Composition}
The final image is obtained by blending our style consistent foreground image generated from FG-Gen and geometrically consistent background generated from BG-Mod.

\subsection{Varying positions, sizes and poses} \label{sec:vary}

By manipulating the input latent codes, our proposed FBC-GAN can generate images with the same background scene and different foreground objects as well as images with the same foreground objects and different background scenes. However, the most unique feature of our FBC-GAN is that it can vary the positions, sizes, and poses of the foreground objects in generated images while preserving the exact identity of the foreground object without additional conditions. This is attributed to the generation of pure foreground objects (zero background information) by our unique generation process. That is, instead of controlling the foreground sizes, positions and poses through latent code, we can directly apply transformations (shifting, flipping, rotation and resizing) to the generated foreground shapes and foreground objects. BG-Mod can modify the backgrounds accordingly to fit the transformed foreground objects. Such unique generation process endows the network with the capbility to generate images that can vary the positions, sizes and poses of the foreground object while preserving the exact identity of the foreground objects without additional conditions required.

\subsection{Training Objective}

The loss $\mathcal{L}_{bg}$ of background generation for BG-Gen follows $\mathcal{L}_{b}$ in FineGAN~\cite{zhao2019image}. 
The generated foreground by FG-Gen and foreground shape by S-Gen are learned via adversarial learning with loss $\mathcal{L}_{fg\_adv}$ and $\mathcal{L}_{s\_adv}$, respectively. To generate more visual details for foreground, we introduce additional feature matching loss $\mathcal{L}_{FM}$ and perceptual loss $\mathcal{L}_{P}$ for foreground generation and set their weights to be 10 ~\cite{wang2018high}. 
Thus, the foreground loss $\mathcal{L}_{fg}$ can be formulated as
\begin{align}\begin{aligned}
\mathcal{L}_{fg}= \mathcal{L}_{fg\_adv} + 10\mathcal{L}_{FM} + 10\mathcal{L}_{P}
\end{aligned}\end{align}

The preliminarily generated image $y$ from $G_{b3}$ is learned via adversarial learning with loss $\mathcal{L}_{img\_adv}$. Inspired by the idea of GAN-CLS~\cite{reed2016generative}, we align $y$ with the generated mask $m_{g}$ by a matching aware discriminator $D_{img\_seg}$ that implicitly discriminate between true image-segmentation pair($x$, $m$) and two sources of fake pairs: ($y$, $m_{g}$) and ($x$, $m_w$). $x$ is the real image, $m$ is its corresponding foreground object mask and $m_w$ is a mismatched foreground object mask from another real image. We therefore formulate the adversarial loss for image and segmentation pair  $\mathcal{L}_{imgseg\_adv}$ by: 
\begin{align}\left.\begin{aligned}
\mathcal{L}_{imgseg\_adv}=\mathop{min}\limits_{G_{b3}}\mathop{max}\limits_{D_{img\_seg}}\mathbb{E}[logD_{img\_seg}(x, m)]\\
+ \mathbb{E}[log(1-D_{img\_seg}(x, m_w)] \\
+ \mathbb{E}[log(1-D_{img\_seg}(y, m_g)] 
\end{aligned}\right.\end{align}

The foreground object shape in $y$ should be the same as the generted shape. This can be restricted by aligning the generated mask $m_{g}$ from $G_{b3}$ with the generated shape $m_i$ from S-Gen . We introduce mean square error loss $\mathcal{L}_{fg\_shape}$ to achieve this alignment. 
\begin{align}\begin{aligned}
\mathcal{L}_{fg\_shape} = ||m_{g}-m_{i}||_2
\end{aligned}\end{align}

To ensure that $G_{b3}$ will not modify background information massively, we design an attention-based background loss between the generated background $y_{bg}$ and the background of $y$, where attention mask $m_a$ is also generated to learn the attention of the reserved background areas. For our generator to reserve as much background information as possible and rationalize the overall image with minimum alternation, the attention mask shall be as close as the reverse of input shape prior $m_i$. The attention-based background loss $\mathcal{L}_{attn\_bg}$ is thus defined by:

\begin{align}\begin{aligned}
  \mathcal{L}_{attn\_bg} = ||m_a*y_{bg} - m_a * y||_2  \\
+ ||m_{a}-(1 - m_{i})||_2
\end{aligned}\end{align}

\smallskip
The overall loss function of the proposed FBC-GAN is:
\begin{align}\begin{aligned}
  \mathcal{L} = \mathcal{L}_{bg} + \mathcal{L}_{fg} + \mathcal{L}_{s\_adv} + \mathcal{L}_{img\_adv}\\
  + \mathcal{L}_{imgseg\_adv}  + \mathcal{L}_{fg\_shape} + \lambda_{2} \mathcal{L}_{attn\_{bg}}
\end{aligned}\end{align}
We empirically set $\lambda_{1}$ as $200$ and $\lambda_{2}$ as $50$.

\begin{figure*}[t] 
\begin{center}
 \includegraphics[width=0.975\linewidth, height=64mm]{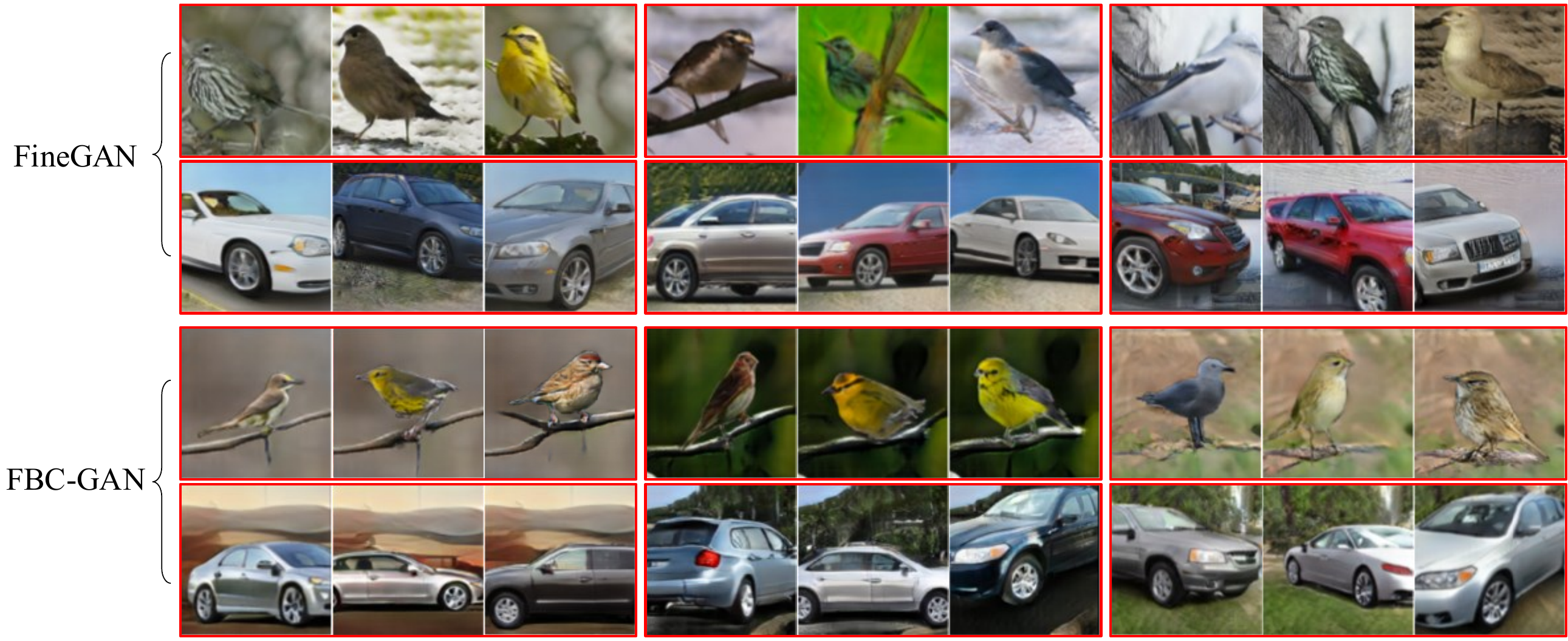}
\end{center}
   \caption{
    Qualitative comparison of FBC-GAN with FineGAN: Images in each red box are a group of images with fixed background latent codes.
   }   
     
\label{fig:same-b-diff-f}
\end{figure*}

\section{Experiments}

\subsection{Datasets and Implementation Details} \label{dataset}

Our network generates image foregrounds and backgrounds independently and reinstate their style and geometry correlations for synthesis realism. Thus, our generated foregrounds and backgrounds can be sampled from the same dataset or from from different datasets. We evaluate such superiority over five different datasets as shown in table~\ref{table:Datasets}.
Each dataset consists of a foreground sub-dataset and a background sub-dataset, 
where foreground sub-dataset contains foreground objects that is used to train FG-Gen and foreground shapes that is use to train S-Gen. And background sub-dataset contains images 
and foreground shapes where foreground shapes are used to guide BG-Mod in generating geometrically compatible background scenes. Original Stanford Cars~\cite{wah2011caltech} dataset and ImageNet~\cite{imagenet_cvpr09}-bird dataset do not contain object shape, thus, we deploy DeepLab-V3~\cite{chen2017rethinking} to obtain the object shape and use VGG19 to screen out well-segmented (classification accuracy of segmented objects higher than 0.5) shapes and their corresponding images and foreground objects. As a result, there left 31979 out of 46800 image sets for ImageNet~\cite{imagenet_cvpr09}-bird and 3196 out of 8144 image sets for Stanford Cars~\cite{wah2011caltech}.

 We use Adam optimizer to train the model  with $\beta_1$ = $0$ and $\beta_2$ = $0.9$ with a fixed learning rate of $0.0002$.

\begin{figure*}[t] 
\begin{center}
   \includegraphics[width=0.975\linewidth, height=64mm]{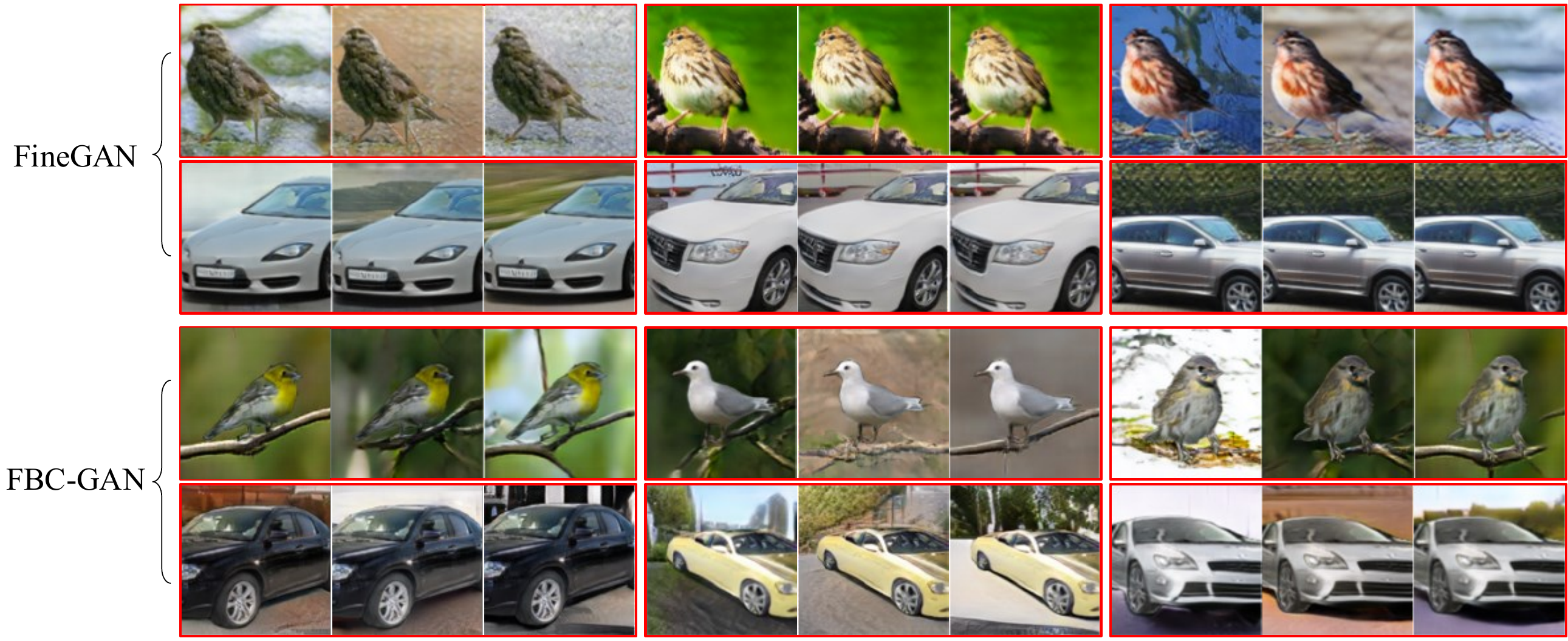}
\end{center}
   \caption{
   Qualitative comparison of FBC-GAN with FineGAN: Images in each red box are a group of images with fixed foreground latent codes. 
   } 
      
\label{fig:same-f-diff-b}
\end{figure*}

\begin{table}
\begin{center}
\scalebox{0.8}{
\begin{tabular}{c c c }
\hline
Name  & Foreground Sub-dataset & Background Sub-dataset \\
\hline\hline
Dataset1 & CUB200~\cite{wah2011caltech}  & CUB200~\cite{wah2011caltech}  \\
\hline
Dataset2 & CUB200~\cite{wah2011caltech}  & ImageNet~\cite{imagenet_cvpr09}-bird \\
\hline
Dataset3 & CUB200~\cite{wah2011caltech}  & Monet Style CUB200~\cite{wah2011caltech} \\
\hline
Dataset4 & Stanford Cars~\cite{krause20133d} & Stanford Cars~\cite{krause20133d}. \\
\hline
Dataset5 & Stanford Cars~\cite{krause20133d} & Monet Style Stanford Cars~\cite{krause20133d} \\
\hline
\end{tabular}}
\end{center}
\caption{Definition of datasets: Each dataset consists of a foreground sub-dataset and a background sub-dataset obtained from the corresponding dataset. ImageNet-bird stands for all datas that are labelled as `bird' in Imagenet. Monet Style CUB200 and  Monet Style Stanford Cars are obtained by transferring nature CUB200 and Stanford Cars images to Monet style images using a pretrained CycleGAN~\cite{CycleGAN2017} model.
}
\label{table:Datasets}
\end{table}

\begin{table} 
\begin{center}
\scalebox{0.8}{

\begin{tabular}{c c c c}
\hline

\multirow{2}*{\minitab{Metrics}} & \multirow{2}*{\minitab{Models}} &  \multicolumn{2}{c}{Scores} \\

&  & Dataset1 & Dataset4 \\ \hline\hline

\multirow{3}*{\minitab{IS \cite{salimans2016improved}}}   
&   StackGAN-V2  &   30.04 $\pm$ 0.5          &   20.66 $\pm$ 0.38 \\
&    FineGAN     &     30.1 $\pm$ 0.64       &      20.34 $\pm$ 0.22 \\
& Ours (FBC-GAN) & \bf{32.2} $\pm$ \bf{0.87}  & \bf{20.89} $\pm$ \bf{0.28} \\ 
\cline{1-4}

\multirow{3}*{\minitab{CIS \cite{huang2018multimodal} }} 
& StackGAN-V2   & 0           & 0   \\
& FineGAN      & 8.2 $\pm$ 0.08  & 6.35 $\pm$ 0.06 \\
& Ours (FBC-GAN) & \bf{31.46} $\pm$ \bf{0.76} & \bf{20.86} $\pm$ \bf{0.37} \\ 
\cline{1-4}

\multirow{3}*{\minitab{LPIPS\cite{zhang2018unreasonable}}} 
& StackGAN-V2 & 0  & 0\\
& FineGAN & 0.38 & 0.32 \\ 
& Ours (FBC-GAN) & \bf{0.42}& \bf{0.54}\\ 
\cline{1-4}

\end{tabular}}
\end{center}
\caption{Quantitative comparison of FBC-GAN with FineGAN and StackGAN-V2 in synthesis fidelity and diversity by using metrics IS, CIS, and LPIPS.
}
\label{table:Overall compraison}
\end{table}

\subsection{Evaluation Metrics} 

We perform quantitative evaluations by using three evaluation metrics. 
The first is Inception Score (IS)~\cite{salimans2016improved} that is commonly used in image synthesis problems. 

The second metric is Conditional IS (CIS) \cite{huang2018multimodal} that defines the inception score conditioned on modes. In our experiments, CIS is evaluated as the inception score of randomly generated images conditioned on the same background. The third metric is Learned Perceptual Image Patch Similarity (LPIPS) \cite{zhang2018unreasonable} that evaluates the distance between image patches. Higher LPIPS means better diversity of generated images.

Frechet Inception Distance (FID)~\cite{heusel2017gans} is another metric to measure the image fidelity. It calculates the distance between feature vectors of real and generated images. However, our model lifts the content correlation between generated foregrounds and backgrounds, making generated images move beyond the distribution of original dataset. Thus, it is not rationale to compare the feature vectors distance between our generated images and original dataset.

In our experiments, we compute IS over 30k randomly generated images for IS. For CIS, we evaluate over 10 groups of randomly generated images, where each group consists of 30k image samples conditioned on the same background. For LPIPS, we follow MUNIT~\cite{huang2018multimodal} to generate 100 groups of images where each group contains 5000 samples conditioned on the same background and randomly selects 19 pairs of images from each group.

\subsection{Quantitative Experiments}
We compare FBC-GAN with FineGAN~\cite{zhao2019image} and StackGAN-V2~\cite{zhang2018stackgan++} quantitatively to demonstrate that our FBC-GAN can generate images with comparable fidelity and higher diversity. FineGAN~\cite{zhao2019image} is the state-of-the-art composition-based GAN for image synthesis, while StackGAN-V2~\cite{zhang2018stackgan++} currently has the best performance on CUB200~\cite{wah2011caltech} and Stanford Cars~\cite{krause20133d} used in our paper. We only conduct quantitative experiments on Dataset1 and Dataset4 as other datasets consist of foreground sub-dataset and a background sub-dataset obtained from different datasets which do not apply to FineGAN and StackGAN-V2. 

For fair comparison, we report results of Finegan without hard negative training as this training technique is only applicable to class conditional models. And when comparing CIS and LPIP, we tie the background and child codes of FineGAN as the style of generated foreground and background are consistent only in this setting. All generated images have the same resolution of $128 \times 128$.

\begin{figure}[t] 
\begin{center}
   \includegraphics[width=0.9\linewidth, height=40.0mm]{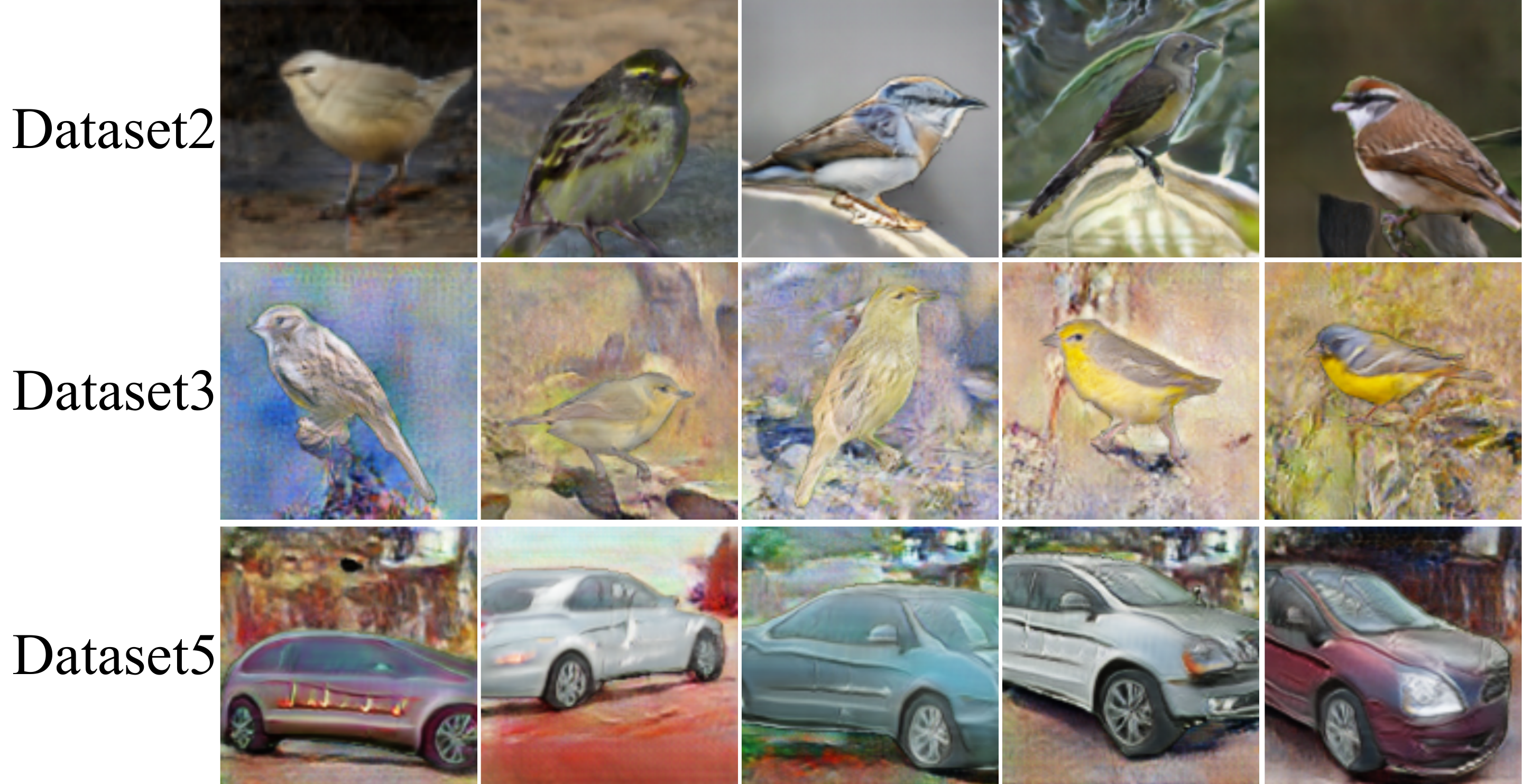}
\end{center}
   \caption{
   Qualitative results for Dataset2, Dataset3 and Dataset5 as defined in Table \ref{table:Datasets}: Each image is finally composed to be style-aligned and geometry-aligned.
   } 
      
\label{fig:d2d3}
\end{figure}

Table~\ref{table:Overall compraison} shows quantitative experimental results. We can observe that FBC-GAN achieves the comparable IS, which means that its generated images have the comparable quality and fidelity. In addition, FBC-GAN achieves the best CIS and LPIPS demonstrating the superior diversity of its generated images as illustrated in Figs.~\ref{fig:same-b-diff-f} and~\ref{fig:same-f-diff-b} (more details to be discussed in the ensuing subsection). CIS and LPIPS for StackGAN-V2 are 0 as it is not able to generate different foreground object for a fixed background. The quantitative experiments show that FBC-GAN is capable of generating high-fidelity and high-diversity images consistently across the three adopted evaluation metrics.

\subsection{Qualitative Experiments} 

We further present qualitative experimental results of FBC-GAN to demonstrate its flexibility and diversity in image synthesis. We also compare FBC-GAN with FineGAN, the state-of-the-art composition-based GAN that similarly tries to disentangle the foreground and background generation processes.

\begin{table*} [t]
\begin{center}
\scalebox{0.9}{
\begin{tabular}{c c c c c c c}
\hline

\multirow{2}*{\minitab{Metrics}} & \multirow{2}*{\minitab{Design Choice}} &  \multicolumn{5}{c}{Scores} \\

& & Dataset1 & Dataset2 &Dataset3 & Dataset4 & Dataset5\\ \hline\hline
\multirow{2}*{\minitab{Style Relevance\cite{zhang2020cross}}} & w/o Style Alignment & 0.254 & 0.261 & 0.230& 0.285 & 0.255\\
& w/ Style Alignment & \bf{0.274} &\bf{0.273}& \bf{0.253} & \bf{0.295} & \bf{0.282} \\
\hline
\multirow{2}*{\minitab{IS~\cite{salimans2016improved}}} & w/o Geometrical Alignment & 30.6 $\pm$ 0.60  & 25.46 $\pm$ 0.46& NA & 20.14 $\pm$ 0.32 & NA \\
& w/ Geometrical Alignment & \bf{32.2} $\pm$ \bf{0.87} &  \bf{26.03} $\pm$ \bf{0.45} & NA & \bf{20.89} $\pm$ \bf{0.28} &  NA \\

\cline{1-7}
\end{tabular}}
\end{center}
\caption{Quantitative ablation study for generated images with and without style alignment using Style Relevance~\cite{zhang2020cross} as well as Quantitative ablation study for generated images with and without geometrical alignment using IS~\cite{salimans2016improved} over five datasets. IS~\cite{salimans2016improved} is not applicable to Dataset3 and Dataset5 as they generate Monet style final images. 
}
\label{table:style relevence}
\end{table*} 

\begin{figure}[t] 
\begin{center}
   \includegraphics[width=0.9\linewidth, height=28mm]{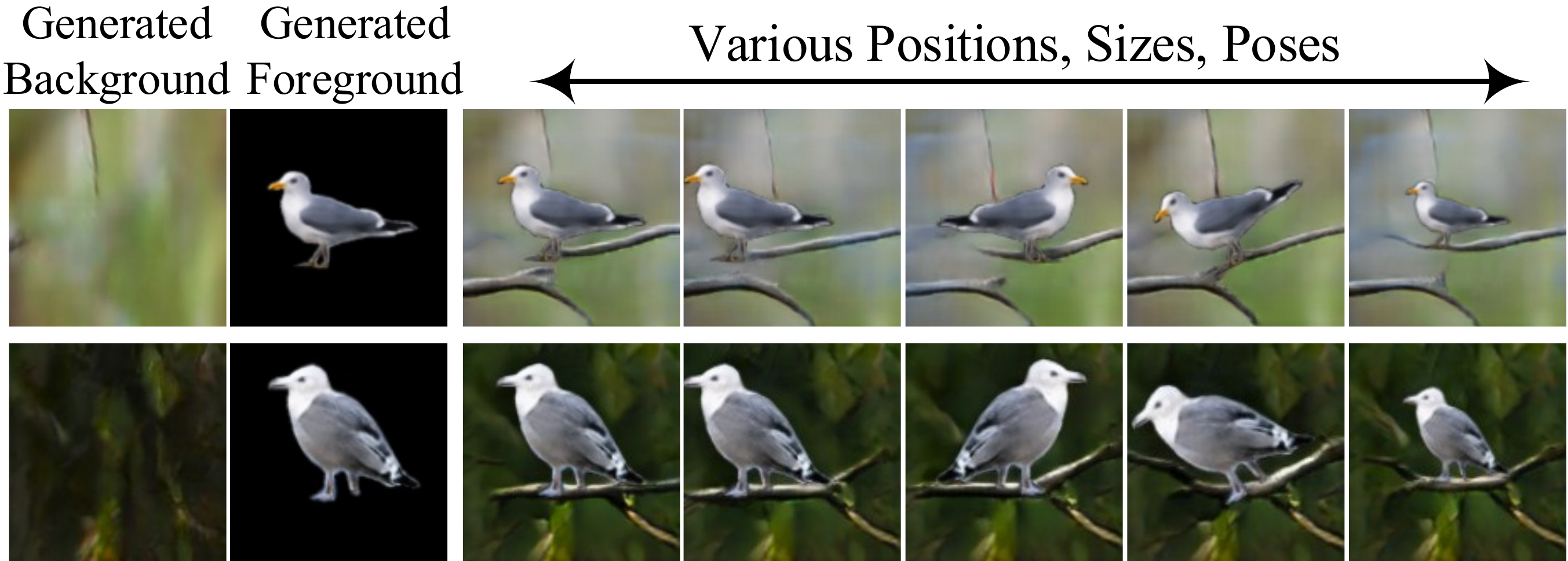}
\end{center}
   \caption{FBC-GAN allows the generated foreground objects to appear at different positions and with different sizes and poses within the same background image. }
\label{fig:same-f-same-b}
\end{figure}

First two rows and last two rows of Fig.~\ref{fig:same-b-diff-f} are generated results for FineGAN and FBC-GAN, respectively. Every three images within a red box are images generated conditioned on the same background. As a comparison, FBC-GAN has superior flexibility in keeping the background unchanged. This is because FBC-GAN generates foreground objects with zero background information and composes them into geometrically compatible backgrounds that only make minimum alterations to originally generated backgrounds from BG-Gen. However, FineGAN tends to generate foregrounds with their own backgrounds, and it will wipe out the originally generated background if generated foregrounds contain too many backgrounds on their own.

Due to the independent generation of foreground objects and background images, FBC-GAN is capable of embedding the same foreground objects into different background images with good consistency in both style and geometry. As Fig.~\ref{fig:same-f-diff-b} shows, FBC-GAN can embed the same foreground object into different background images realistically. As a comparison, FineGAN can disentangle the foreground and background generation processes as well. On the other hand, its generated foreground can occlude the whole background with similar reason as described for Fig.~\ref{fig:same-b-diff-f}. As a result, all generated images will be the same. Moreover, it is more rationale if the same foreground object has its correlated style appearing at different background scenes. Our FBC-GAN is capable of achieving such a requirement (e.g., the first image set generated by FBC-GAN shows foreground object is dimmer if the background scene is dimmer) while generated foreground objects do not change accordingly using FineGAN. 

Fig.\ref{fig:same-b-diff-f} and \ref{fig:same-f-diff-b} demonstrate the generation results of Dataset1 and Dataset4 using our FBC-GAN. Dataset1 and Dataset4 are datasets where foreground sub-dataset and background sub-dataset are sampled from the same dataset, and to demonstrate the flexibility of our FBC-GAN, Fig.~\ref{fig:d2d3} shows qualitative experimental results of another three datasets where each dataset consists of a foreground sub-dataset and a background sub-dataset obtained from different datasets. We find that although foregrounds and backgrounds are from different datasets, the final generated images are still visual realistic with style alignment and geometric alignment design in our FBC-GAN.

\begin{figure}[t] 
\begin{center}
   \includegraphics[width=0.975\linewidth, height=50mm]{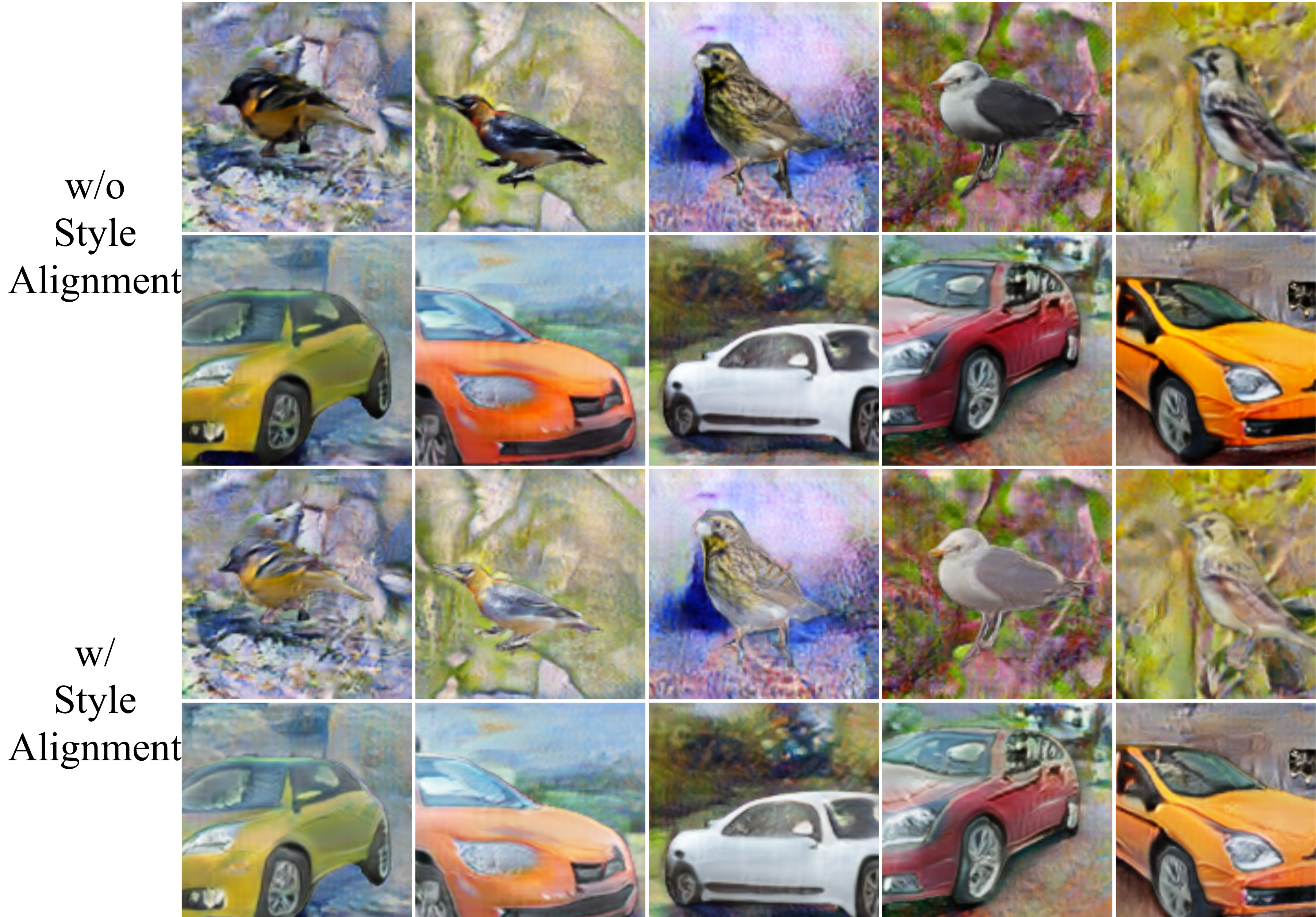}
\end{center}
   \caption{Qualitative ablation study for style alignment: Upper two rows and lower two rows show the composed images without and with our style alignment, respectively. 
   }
\label{fig:Ablation study for style alignment}
\end{figure}

Further, FBC-GAN has a unique feature that without requiring additional conditions, it allows the same generated foreground objects to appear at different positions with different sizes and poses within the same background as shown in Fig.~\ref{fig:same-f-same-b}. This is largely attributed to our unique network generation design (details discussed in section \ref{sec:vary}).

This unique feature allows FBC-GAN to generate more images which further improves the generation diversity greatly.

\subsection{Ablation Study}

\begin{figure}[t] 
\begin{center}
  \includegraphics[width=0.975\linewidth, height=50mm]{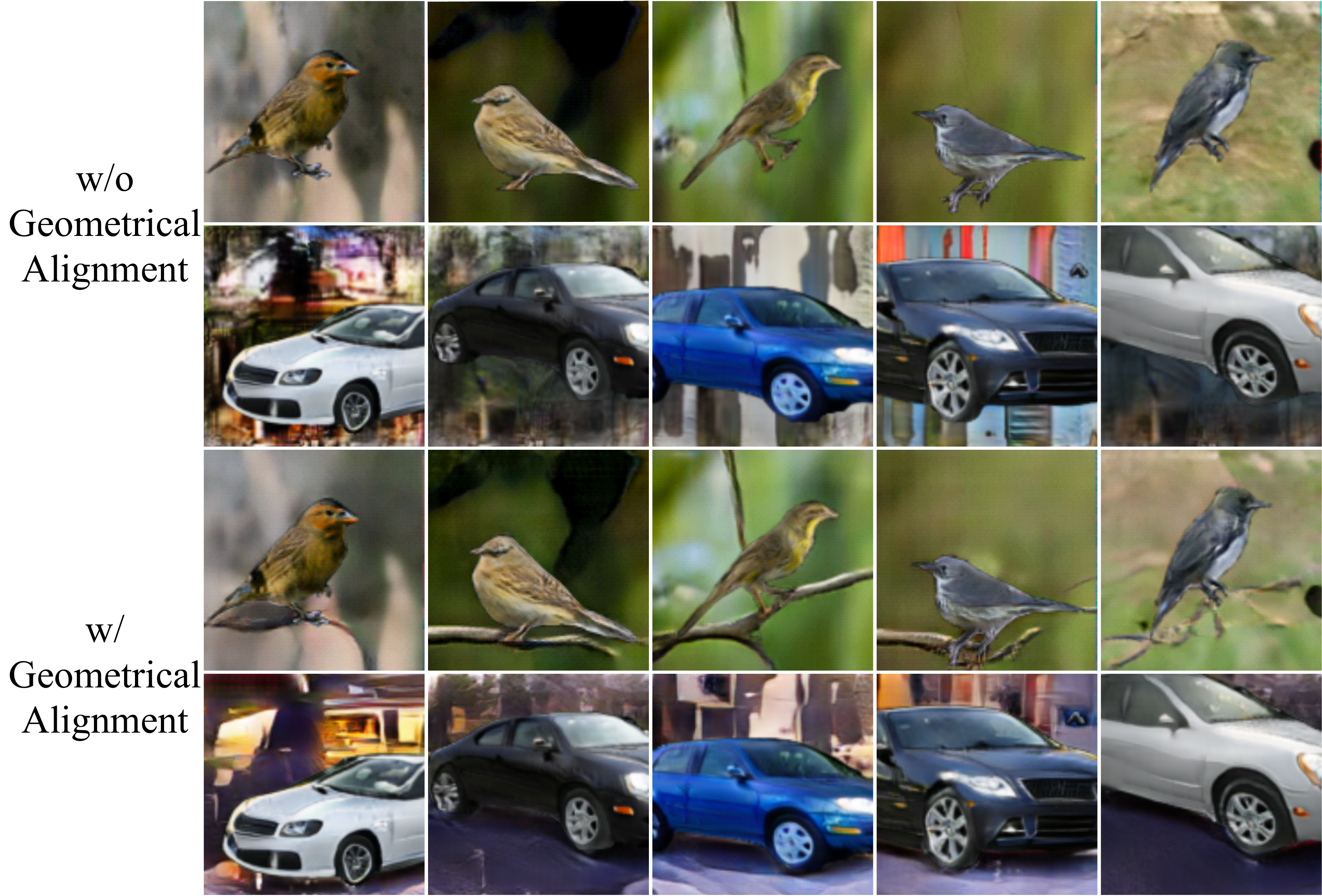}
\end{center}
  \caption{Qualitative ablation study for geometrical alignment: Upper two rows and lower two rows show the composed images without and with our geometry alignment, respectively.
  }
\label{fig: Ablation study for geometrical alignment}
\end{figure}

We perform ablation studies quantitatively and qualitatively to demonstrate the effectiveness of our style alignment and geometry alignment in image synthesis. We use Style Relevance  \cite{zhang2020cross} to show the style compatibility between generated foreground and background (evaluate over 30k generated foregrounds and backgrounds). First block in Table \ref{table:style relevence} shows that with our designed style consistency mechanism, style aligned foregrounds consistently have better style compatibility over the five datasets. Demonstrated images in Fig.~\ref{fig:Ablation study for style alignment} are composed of foreground objects sampled from nature style dataset and background scenes sampled from Monet style dataset, it is intuitive that style aligned images have more compatible foreground and background, in brightness, texture, etc. Second block in Table \ref{table:style relevence} shows quantitatively that generated images with our geometry consistency mechanism have higher quality consistently.
And Fig.~\ref{fig: Ablation study for geometrical alignment} further proves that our geometry alignment mechanism  rationalizes the generated foreground and background geometrically, and the composed images has poor geometry fidelity without our geometric alignment.

\section{Conclusion}

This paper presents a novel Foreground-Background Composition GAN (FBC-GAN) that positions foreground and background generation as two independent processes and offers superior flexibility and diversity in image generation.
FBC-GAN consists of a Foreground Generator (FG-Gen) and Background Generator (BG-Gen) to generate foreground and background concurrently and independently. It adapts the idea of AdaIN \cite{huang2017arbitrary} for style alignment and it deploys Shape Generator (S-Gen) and Background Modifier (BG-Mod) to bridge the geometry between the generated foreground and background. 
Due to the novel generation process, FBC-GAN lifts the content correlation between the generated foreground and background, 
allowing generations with the same background but different foreground, the same foreground but different background, the same foreground and background but different foreground object positions, poses, etc. Additionally, it also allows to generate `new information' by enabling image synthesis with foreground and background sampled from different datasets. Extensive experiments show the superiority of our synthesis network.

{\small
\bibliographystyle{ieee_fullname}
\bibliography{egbib}
}

\newpage

\begin{figure*}[t] 
\begin{center}
\includegraphics[width=1\linewidth, height=90mm]{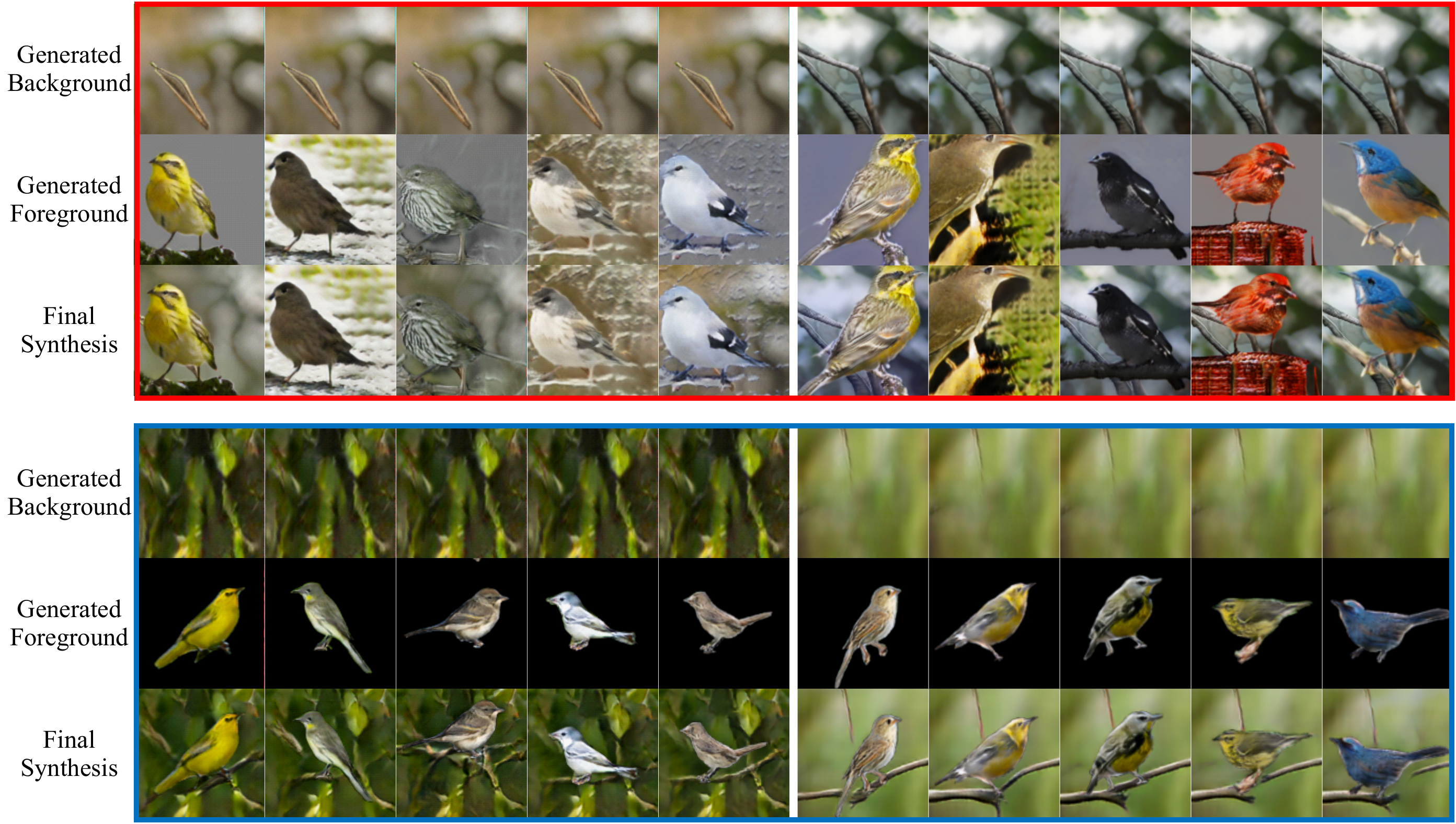}
\end{center}
   \caption{
    Qualitative comparison of FBC-GAN with FineGAN over CUB200 when background latent codes are fixed (for Dataset1): Images in red and blue boxes are generated by FineGAN and our proposed FBC-GAN, respectively. Our FBC-GAN has better geometry plausibility by adjusting its background with respect to the foreground (e.g. the tree branches).
   }   
\label{fig:same-b-diff-f_cub_app}
\end{figure*} 

\begin{figure*}[t] 
\begin{center}
\includegraphics[width=1\linewidth, height=90mm]{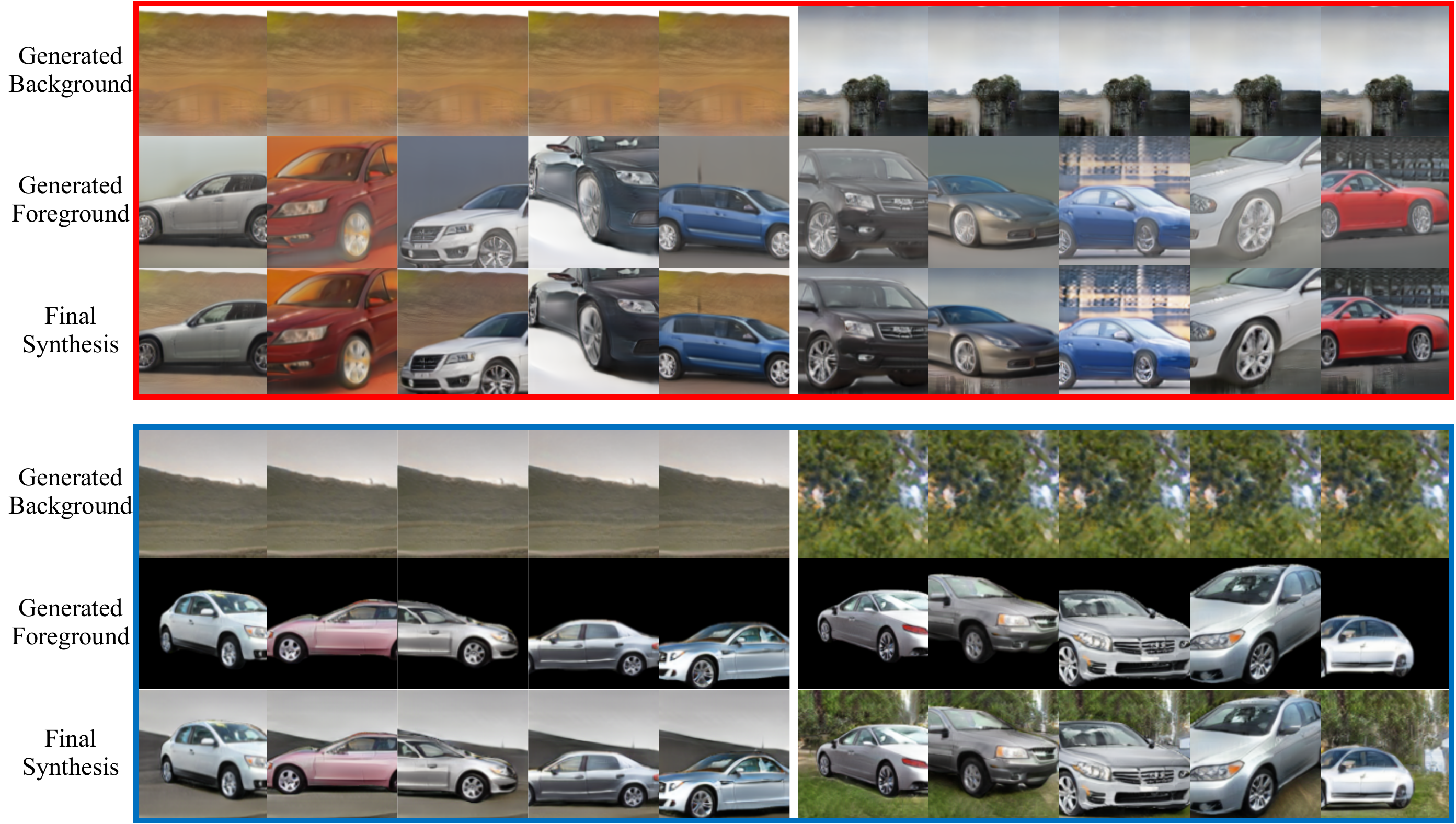}
\end{center}
   \caption{
    Qualitative comparison of FBC-GAN with FineGAN over Stanford Cars when background latent codes are fixed (for Dataset1): Images in red and blue boxes are generated by FineGAN and our proposed FBC-GAN, respectively.
   }   
\label{fig:same-b-diff-f_car_app}
\end{figure*} 

\begin{figure*}[t] 
\begin{center}
\includegraphics[width=1\linewidth, height=90mm]{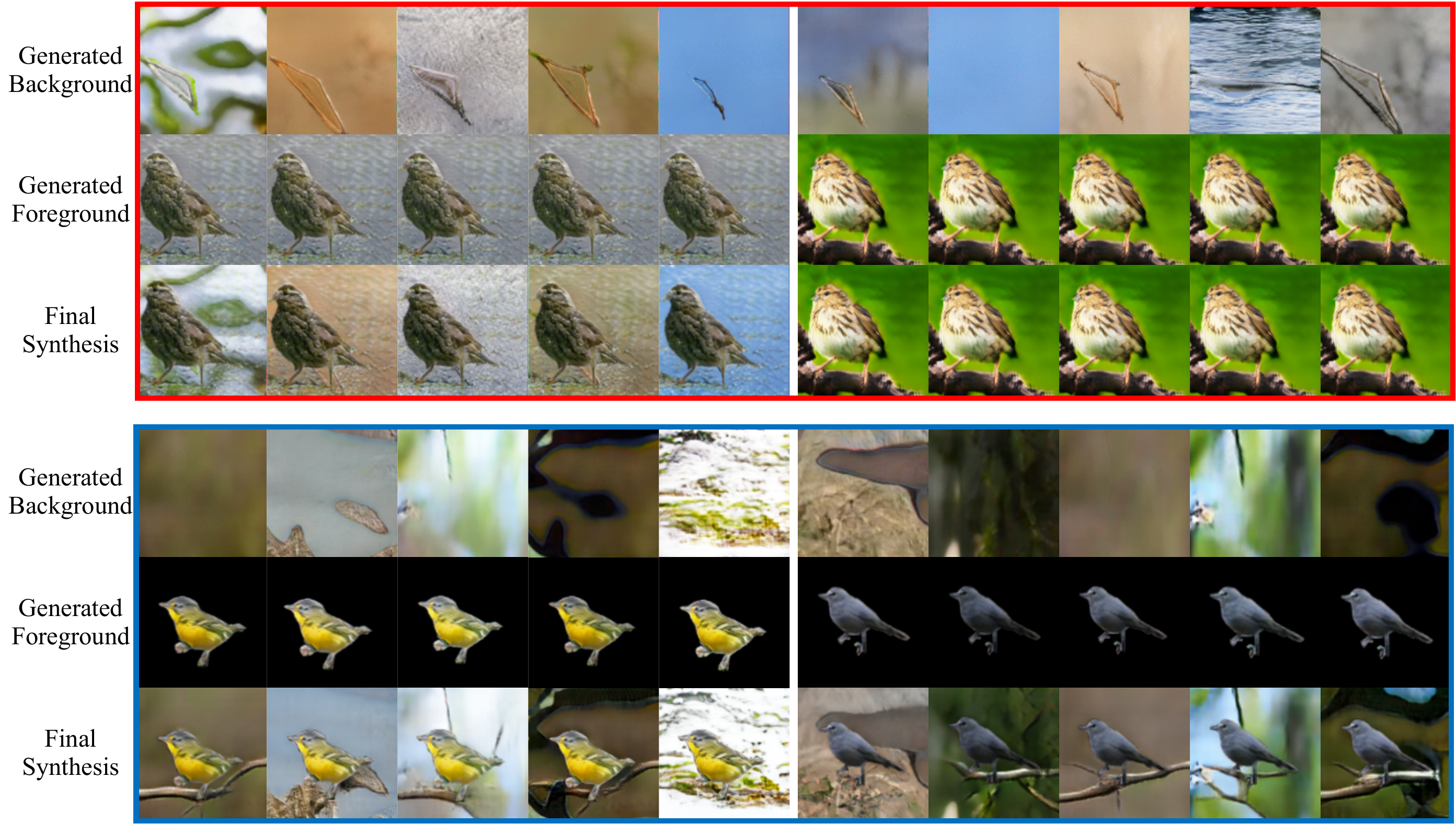}
\end{center}
   \caption{
    Qualitative comparison of FBC-GAN with FineGAN over CUB200 when foreground latent codes are fixed (for Dataset4): Images in red and blue boxes are generated by FineGAN and our proposed FBC-GAN, respectively. Our FBC-GAN has better geometry plausibility by adjusting its background with respect to the foreground (e.g. the tree branches).
   }   
\label{fig:same-f-diff-b_cub_app}
\end{figure*}

\begin{figure*}[t] 
\begin{center}
\includegraphics[width=1\linewidth, height=90mm]{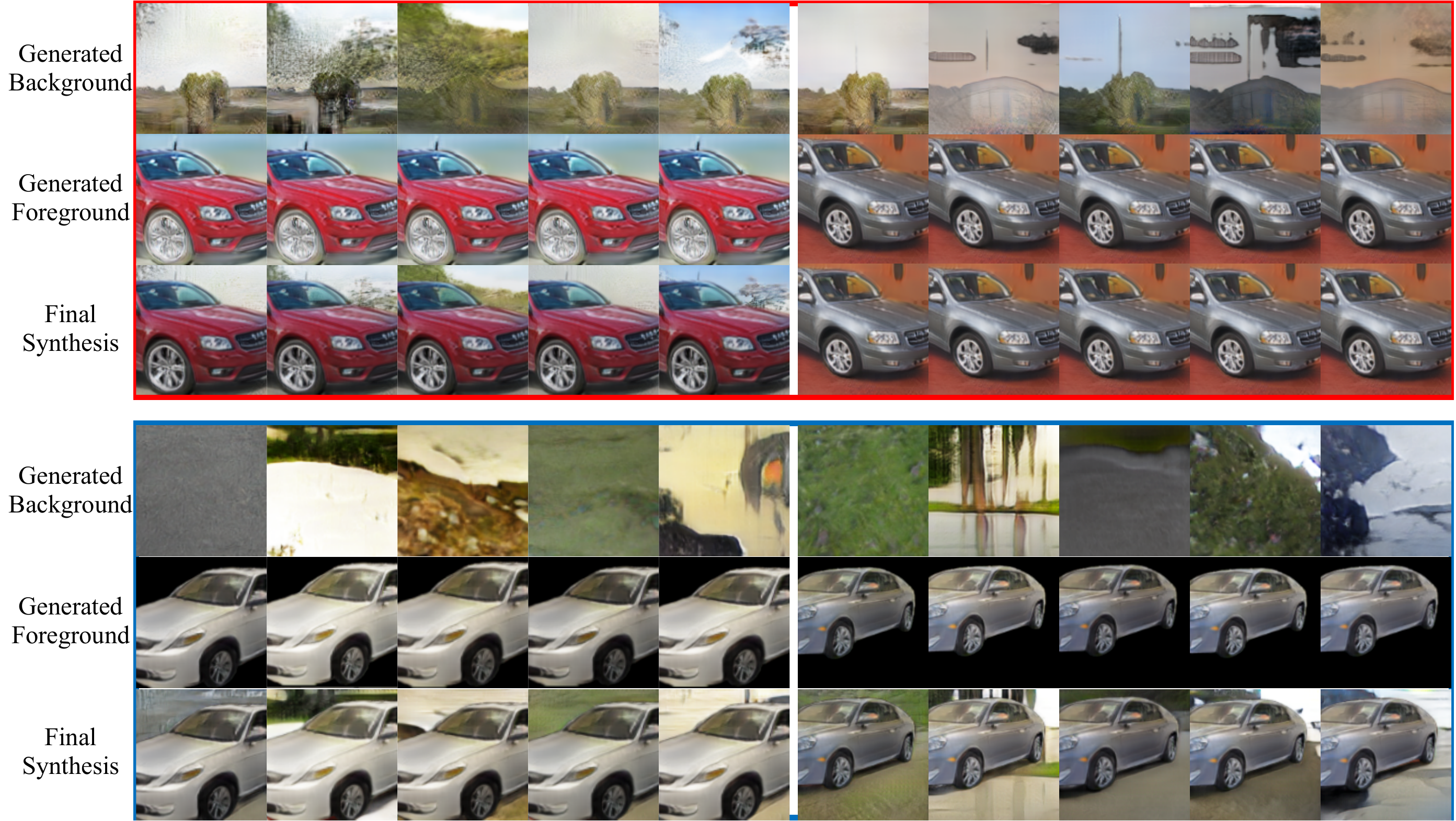}
\end{center}
   \caption{
    Qualitative comparison of FBC-GAN with FineGAN over Stanford Cars when foreground latent codes are fixed (for Dataset4): Images in red and blue boxes are generated by FineGAN and our proposed FBC-GAN, respectively.
   }   
\label{fig:same-f-diff-b_car_app}
\end{figure*} 

\begin{figure*}[t] 
\begin{center}
\includegraphics[width=1\linewidth, height=110mm]{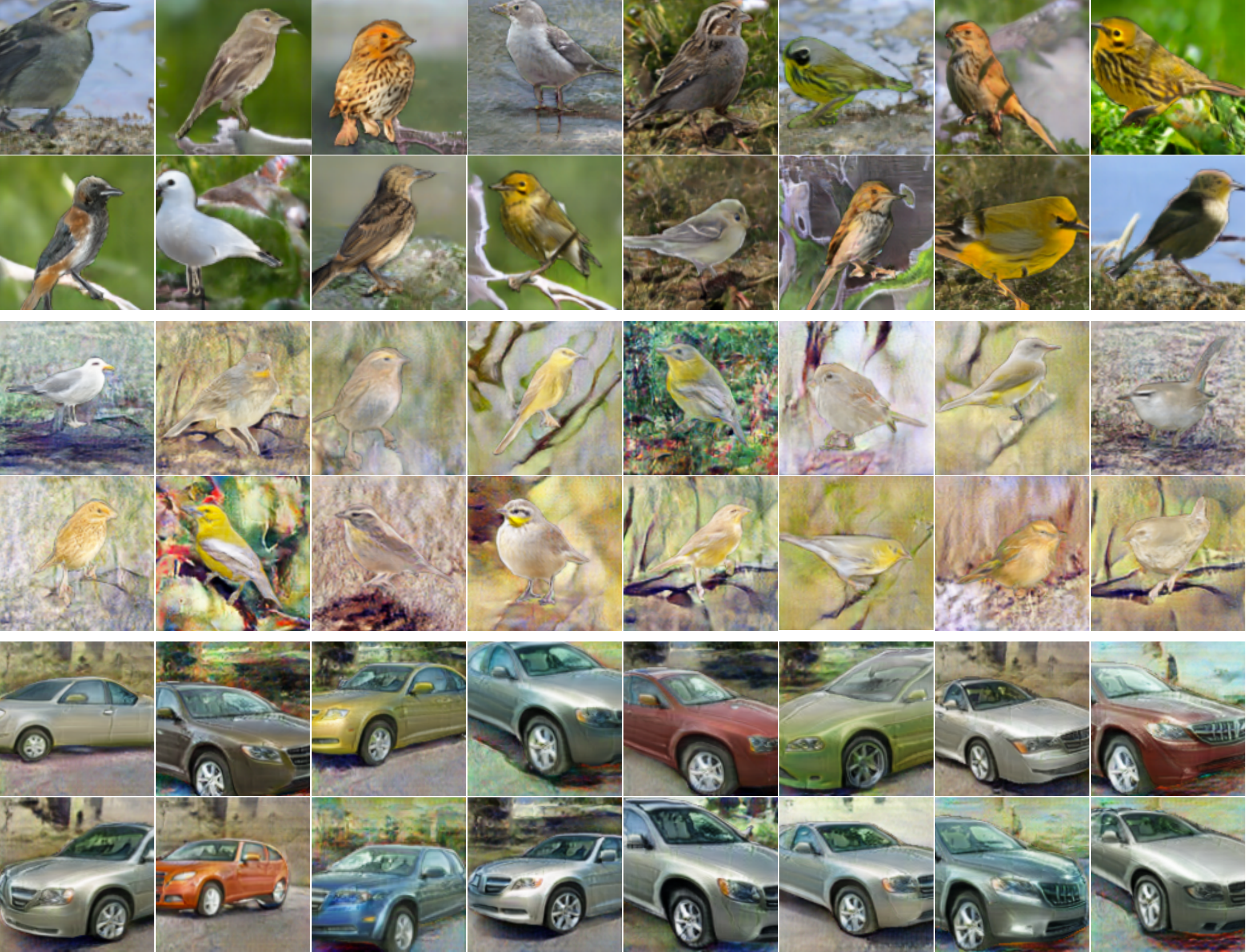}
\end{center}
   \caption{
    Qualitative experimental results for Dataset2, Dataset3 and Dataset5: Rows 1-2 show generated images for Dataset2, Row 3-4 show generated images for Dataset3, and Rows 5-6 show generated images for Dataset5.
   }   
\label{fig:dataset2}
\end{figure*}

\begin{figure*}[t] 
\begin{center}
\includegraphics[width=1\linewidth, height=66mm]{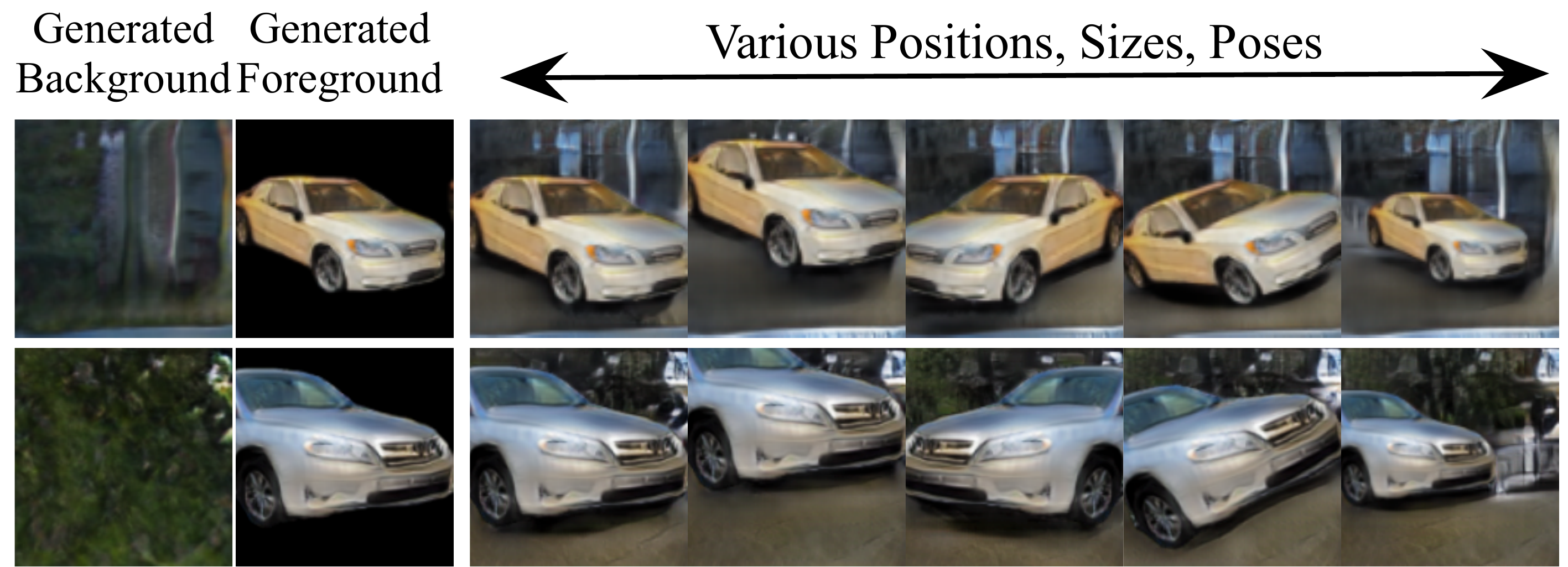}
\end{center}
   \caption{
   FBC-GAN allows the generated foreground objects to appear at different positions and with different sizes and poses within the same background image.
   }   
\label{fig:SFSB_car}
\end{figure*}

\section*{Appendix}

We present additional experimental results over multiple data settings to demonstrate the superior flexibility and diversity as well as generality of our proposed FBC-GAN.

\medskip

Figs.~\ref{fig:same-b-diff-f_cub_app} and~\ref{fig:same-b-diff-f_car_app} show qualitative comparisons of FineGAN and our FBC-GAN when background latent codes are fixed over datasets CUB200 and Stanford Cars. To demonstrate detailed differences between FineGAN and FBC-GAN, we also compare the generated foregrounds and backgrounds. 
With background latent codes fixed, the generated image backgrounds should ideally be the same. However, FineGAN tends to generate foregrounds with their own backgrounds, and it will wipe out the originally generated background if generated foreground images has too much background on their own. Therefore, FineGAN cannot guarantee fixed background. As a comparison, our FBC-GAN ensures that all finally generated images have the same background as long as the background latent codes are fixed. This is because FBC-GAN generates foreground objects with no background, and then composes them into geometrically compatible backgrounds with minimal alterations of the originally generated background. Such alternation without substantial change in originally generated background merely serves to avoid geometrical inconsistency like a bird standing in the air.

In addition, Figs.~\ref{fig:same-f-diff-b_cub_app} and~\ref{fig:same-f-diff-b_car_app} show qualitative comparison with FineGAN with foreground latent codes fixed over CUB200 dataset and Stanford Cars dataset, respectively. We can spot the same issue for FineGAN: due to incomplete foreground-background disentangle, FineGAN generated foregrounds could wipe out the originally generated backgrounds. With changing background latent codes, FineGAN often fails to generate different backgrounds. 

In addition, it is natural that the style of foreground object(s) should be consistent with the corresponding background.
With this in mind, our proposed FBC-GAN can adjust the styles of foreground objects to be compatible with the separately generated background, making the final synthesized image more visually realistic. However, FineGAN lacks this desirable property.

We defined 5 datasets in this work. We further present more illustrations in Figs. \ref{fig:same-b-diff-f_cub_app} and \ref{fig:same-f-diff-b_cub_app} for Dataset1, and Figs. \ref{fig:same-b-diff-f_car_app} and \ref{fig:same-f-diff-b_car_app} for Dataset4. In addition, Fig.~\ref{fig:dataset2} shows several generation samples for Dataset2, Dataset3 and Dataset5. The successful generation over these 5 datasets demonstrates the generality of our proposed FBC-GAN. Additionally, it shows that our FBC-GAN can sample foregrounds and backgrounds from either same dataset or different datasets. This feature lays the foundation for flexible and diverse image generation with our FBC-GAN.

Another feature of FBC-GAN is that it can generate the same foreground object and background scene with different object positions, sizes, and poses without additional conditions. We achieve this feature by applying transformations (shifting, flipping, rotation, and resizing) directly to the generated foreground object and the generated shape. With background scenes modified accordingly, we can obtain visually realistic generation with the same foreground object and background scene. We demonstrated this feature qualitatively over CUB200 in our submitted manuscript. In this appendix, we further demonstrate this feature over the Stanford Cars as illustrated in Fig.~\ref{fig:SFSB_car}.

\end{document}